\documentclass[12pt,conference]{IEEEtran}
\IEEEoverridecommandlockouts
\usepackage{cite}
\usepackage{amsmath,amssymb,amsfonts}
\usepackage{algorithmic}
\usepackage{graphicx}
\usepackage{textcomp}
\usepackage{CJKutf8}
\usepackage{amssymb}
\usepackage{amsmath}
\DeclareMathOperator*{\argmax}{arg\,max}

\usepackage{url}
\def\BibTeX{{\rm B\kern-.05em{\sc i\kern-.025em b}\kern-.08em
    T\kern-.1667em\lower.7ex\hbox{E}\kern-.125emX}}
\begin{document}

\title{Dual Long Short-Term Memory Networks for Sub-Character Representation Learning}

\author{Han He$^1$, Lei Wu$^2$, Xiaokun Yang$^3$, Hua Yan$^4$  \\ 
Zhimin Gao$^5$, Yi Feng$^6$, George Townsend$^7$ \\
{\small$^{1,2,3,4}$University of Houston-Clear Lake, U.S.A}\\
{\small$^5$University of Houston, U.S.A} {\small$^{6,7}$Alsoma University, Canada} \\
{\small \tt $^{1,2,3,4}$\{heh1996,wul,yangxia,yan\}@uhcl.edu}\\
{\small \tt $^5$zgao5@uh.edu, $^{6,7}$\{feng,townsend\}@algomau.ca}
}

\maketitle

\begin{abstract}
Characters have commonly been regarded as the minimal processing unit in Natural Language Processing (NLP). But many non-latin languages have hieroglyphic writing systems, involving a big alphabet with thousands or millions of characters. Each character is composed of even smaller parts, which are often ignored by the previous work. In this paper, we propose a novel architecture employing two stacked Long Short-Term Memory Networks (LSTMs) to learn sub-character level representation and capture deeper level of semantic meanings. To build a concrete study and substantiate the efficiency of our neural architecture, we take Chinese Word Segmentation as a research case example. Among those languages, Chinese is a typical case, for which every character contains several components called radicals. Our networks employ a shared radical level embedding to solve both Simplified and Traditional Chinese Word Segmentation, without extra Traditional to Simplified Chinese conversion, in such a highly end-to-end way the word segmentation can be significantly simplified compared to the previous work. Radical level embeddings can also capture deeper semantic meaning below character level and improve the system performance of learning. By tying radical and character embeddings together, the parameter count is reduced whereas semantic knowledge is shared and transferred between two levels, boosting the performance largely. On 3 out of 4 Bakeoff 2005 datasets, our method surpassed state-of-the-art results by up to $0.4\%$. Our results are reproducible, source codes and corpora are available on GitHub\footnote{\url{
https://github.com/hankcs/sub-character-cws}
}.

\end{abstract}

\begin{IEEEkeywords}
AI Algorithms and Applications, Deep Learning, Machine Learning Algorithms, Natural Language Processing, Neural Networks, Pattern Recognition
\end{IEEEkeywords}

\section{Introduction}
Unlike English, the alphabet in many non-latin languages is often big and complex. In those hieroglyphic writing systems, every character can be decomposed into smaller parts or sub-characters, and each part has special meanings. But existing methods often follow common processing steps in latin flavor \cite{mikolov2010recurrent,Mikolov:2013wc,2016arXiv160704606B,kim2016character,pinter2017mimicking}, and treat character as the minimal processing unit, leading to a neglecting of information inside non-latin characters. Early work exploiting sub-character information usually treat it as a separate level from character \cite{Sun:2014jn,Li:2015td,Shi:2015vx,Dong:2016bl}, ignoring the language phenomenon that some of those sub-characters themselves are often used as normal characters. From this phenomenon, we gained a new motivation to design a novel neural network architecture for learning character and sub-character representation jointly.

\begin{table}
\begin{center}
\caption{\label{font-table} Illustration of semantic component (Sem.) and phonetic component (Pho.) in Simplified Chinese (SC) and Traditional Chinese (TC). }
\begin{tabular}{|c|cc|c|cc|}
\hline
SC&Sem.&Pho.&TC&Sem.&Pho.\\
\hline
\begin{CJK}{UTF8}{gbsn}鲤\end{CJK}&\begin{CJK}{UTF8}{gbsn}鱼\end{CJK}&\begin{CJK}{UTF8}{gbsn}里\end{CJK}&\begin{CJK}{UTF8}{bsmi}鯉\end{CJK}&\begin{CJK}{UTF8}{bsmi}魚\end{CJK}&\begin{CJK}{UTF8}{bsmi}里\end{CJK}\\
\hline
\begin{CJK}{UTF8}{gbsn}鲢\end{CJK}&\begin{CJK}{UTF8}{gbsn}鱼\end{CJK}&\begin{CJK}{UTF8}{gbsn}连\end{CJK}&\begin{CJK}{UTF8}{bsmi}鰱\end{CJK}&\begin{CJK}{UTF8}{bsmi}魚\end{CJK}&\begin{CJK}{UTF8}{bsmi}連\end{CJK}\\
\hline
\begin{CJK}{UTF8}{gbsn}河\end{CJK}&\begin{CJK}{UTF8}{gbsn}水\end{CJK}&\begin{CJK}{UTF8}{gbsn}可\end{CJK}&\begin{CJK}{UTF8}{bsmi}河\end{CJK}&\begin{CJK}{UTF8}{bsmi}水\end{CJK}&\begin{CJK}{UTF8}{bsmi}可\end{CJK}\\
\hline
\begin{CJK}{UTF8}{gbsn}沟\end{CJK}&\begin{CJK}{UTF8}{gbsn}水\end{CJK}&\begin{CJK}{UTF8}{gbsn}勾\end{CJK}&\begin{CJK}{UTF8}{bsmi}溝\end{CJK}&\begin{CJK}{UTF8}{bsmi}水\end{CJK}&\begin{CJK}{UTF8}{bsmi}冓\end{CJK}\\
\hline
\begin{CJK}{UTF8}{gbsn}捞\end{CJK}&\begin{CJK}{UTF8}{gbsn}手\end{CJK}&\begin{CJK}{UTF8}{gbsn}劳\end{CJK}&\begin{CJK}{UTF8}{bsmi}撈\end{CJK}&\begin{CJK}{UTF8}{bsmi}手\end{CJK}&\begin{CJK}{UTF8}{bsmi}勞\end{CJK}\\
\hline
\begin{CJK}{UTF8}{gbsn}捡\end{CJK}&\begin{CJK}{UTF8}{gbsn}手\end{CJK}&\begin{CJK}{UTF8}{gbsn}佥\end{CJK}&\begin{CJK}{UTF8}{bsmi}撿\end{CJK}&\begin{CJK}{UTF8}{bsmi}手\end{CJK}&\begin{CJK}{UTF8}{bsmi}僉\end{CJK}\\
\hline
\end{tabular}
\vspace{-20pt}
\end{center}
\end{table}

In linguists' view, Chinese writing system is such a highly hieroglyphic language, and it has a long history of character compositionality. Every Chinese character has several radicals (\begin{CJK}{UTF8}{gbsn}“部首”\end{CJK} in Chinese), which serves as semantic component for encoding meaning, or phonetic component for representing pronouciation. For instance, we listed radicals of several Simplified and Traditional Chinese characters in Table \ref{font-table}.
Chinese characters with same semantic component are closely correlated in semantic. As shown above, carp (\begin{CJK}{UTF8}{gbsn}鲤\end{CJK}) and silverfish (\begin{CJK}{UTF8}{gbsn}鲢\end{CJK}) are both fish (\begin{CJK}{UTF8}{gbsn}鱼\end{CJK}). River (\begin{CJK}{UTF8}{gbsn}河\end{CJK}) and gully (\begin{CJK}{UTF8}{gbsn}沟\end{CJK}) are all filled with water (\begin{CJK}{UTF8}{gbsn}水\end{CJK}). To catch (\begin{CJK}{UTF8}{gbsn}捞\end{CJK}) or to pick up (\begin{CJK}{UTF8}{gbsn}捡\end{CJK}) a fish, one needs to use hands (\begin{CJK}{UTF8}{gbsn}手\end{CJK}).
To exploit those semantic meanings under character embedding level, radical embedding emerged since 2014 \cite{Sun:2014jn,Shi:2015vx,Mikolov:2013uz,Dong:2016bl}. These early work treated sub-character and character as two separate levels, omitting that they can actually be unified as single minimal processing unit in language model. Instead of ignoring linguistic knowledge, we respect the divergence of human language, and propose a novel joint learning framework for both character and sub-character representations.

To verify the efficiency of our jointly learnt representations, we conducted extensive experiments on the Chinese Word Segmentation (CWS) task. As those languages often don't have explicit delimiters between words, making it hard to perform later NLP tasks like Information Retrieval or Question Answering. Chinese language is such a typical non-segmented language, which means unlike English language having spaces between every word, Chinese has no explicit word delimiters. Therefore, Chinese Word Segmentation is a preliminary pre-processing step for later Chinese language process tasks. Recently with the rapid rise of deep learning, neural word segmentation approaches arose to reduce efforts in feature engineering \cite{Zheng:2013wj,Collobert:2011tk,Pei:2014vx,Chen:2015wa,Cai:2016tg,2017arXiv170407047C}.

In this paper, we propose a novel model to dive deeper into character embeddings. In our framework, Simplified Chinese and Traditional Chinese corpora are unified via radical embedding, growing an end-to-end model. Every character is converted to a sequence of radicals with its original form. Character embeddings and radical embeddings are pretrained jointly in Bojanowski et al.  \cite{2016arXiv160704606B}'s subword aware method. Finally, we conducted various experiments on corpora from SIGHAN bakeoff 2005. Results showed that our jointly learnt character embedding outperforms conventional character embedding training methods. Our models can improve performance by transfer learning between characters and radicals. The final scores surpassed previous work, and 3 out of 4 even surpassed previous preprocessing-heavy state-of-the-art learning work.

More specifically, the contributions of this paper could be summarized as:
\begin{itemize}
\item Explored a novel sub-character aware neural architecture and unified character and sub-character as one same level embedding.
\item Released the first full Chinese character-radical conversion corpus along with pre-trained embeddings, which can be easily applied on other NLP tasks. Our codes and corpora are freely available for the public.
\end{itemize}

\section{Related Work}

In this section, we review the previous work from 2 directions -- radical embedding and Chinese Word Segmentation.

\subsection{Radical Embedding}

To leverage the semantic meaning inside Chinese characters, Sun et al.\cite{Sun:2014jn} inaugurated radical information to enrich character embedding via softmax classification layer. In similar way, Li et al.\cite{Li:2015td} proposed charCBOW model taking concatenation of the character-level and component-level context embeddings as input. Making networks deeper, Shi et al.\cite{Shi:2015vx} proposed a deep CNN on top of radical embedding pre-trained via CBOW. Instead of utilizing CNNs, following Lample et al.\cite{Lample:2016vz}, Dong et al.\cite{Dong:2016bl} used two level LSTMs taking character embedding and radical embedding as input respectively.

Our work is closely related to Dong et al.\cite{Dong:2016bl}, but there are two major differences. In pre-training phase, their character embeddings were pre-trained separately, by utilizing conventional word2vec package, and the radical embeddings are randomly initialized. While we considered radical units as sub-characters (parts of one character) and trained the two level embeddings jointly, following Bojanowski et al. \cite{2016arXiv160704606B}'s approach. In training and testing phases, our two-level embeddings are tied up and unified as the sole minimal input unit of Chinese language.

\subsection{Chinese Word Segmentation}

Chinese Word Segmentation has been a well-known NLP task for decades\cite{Huang2007Chinese}. After pioneer Xue et al.\cite{Xue:2003ti} transformed CWS into a character-based tagging problem, Peng et al. \cite{peng2004chinese} adopted CRF as the sequence labeling model and showed its effectiveness. Following these pioneers, later sequence labeling based work \cite{Tseng2005A,Zhao:2006vi,Zhao2010A,sun2012fast} was proposed. Recent neural models \cite{Zheng:2013wj,Qi:2014uh,Pei:2014vx,Chen:2015wa,Dong:2016bl,2017arXiv170407556C} also followed this sequence labeling fashion.

Our model is based on Bi-LSTM with CRF as top layer. Unlike previous approaches, the inputs to our model are both character and radical embeddings. Furthermore, we explored which embedding level is more tailored for Chinese language, either using both embeddings together, or even tying them up.

\section{Joint Learning for Character Embedding and Radical Embedding}

Previous work treated character and radical as two different levels, used them separately or used one to enhance the other. Although radicals are components of a character (belonging to a lower level), they can actually be learnt jointly. It is linguistically more reasonable to put radical embeddings and character embeddings in exactly the same vector space. We propose to train character vector representation being aware of its internal structure of radicals.

\subsection{Character Decomposition}

Every character can be decomposed into a list of radicals or components. To maintain character information in radical list, we simply add the raw form of character to its radical list. Taking the linguistic knowledge that semantic component contains richest meaning of one character into consideration, we append the semantic component to the end of its radical list, hence to make the semantic component appear more than once.

Formally, denote $c$ as a character, $r$ as a radical, $\mathcal{L}_c = \left[r_1, r_2 \cdots r_n \right]$ as the original radical list of $c$. Let $r_s \in \mathcal{L}_c$ be the semantic component of $c$. Our decomposition of $c$ will be:

\begin{equation}
\mathcal{R}_c=\left[c, r_1, r_2 \cdots r_n, r_s\right]	
\end{equation}

\subsection{General Continuous Skip-Gram (SG) Model}

Take a brief review of the continuous skip-gram model introduced by Mikolov et al.\cite{Mikolov:2013uz}, applied in character representation learning.

Given an alphabet, target is to learn a vectorial representation $\mathbf{v}_{c}$ for each character $c$. Let $c_1, ..., c_T$ be a large-scale corpus represented as a sequence of characters, the objective function of the skipgram model is to maximize the log-likelihood of correct prediction.
The probability of a context character $c_y$ given $c_x$ is computed by a scoring function~$s$ which maps character and context to scores in~$\mathbb{R}$.

The general SG model ignores the radical structure of characters, we propose a different scoring function $s$, in order to capture radical information.

Let all radicals form an alphabet of size $R$. Given a character $c$ and the radical list $\mathcal{R}_c \subset \{1, \dots, R \}$ of $c$, a vector representation $\mathbf{z}_r$ is associated to each radical $r$.
Then a character is represented by the sum of the vector representations of its radicals.
Thus the new scoring function will be:
\begin{equation}
s(c_x, c_y) = \sum_{r \in \mathcal{R}_{c_x}} \mathbf{z}_r^\top \mathbf{v}_{c_y}.
\end{equation}

This simple model allows learning the representations of characters and radicals jointly.

\section{Radical Aware Neural Architectures for General Chinese Word Segmentation}

Once character and radical representations are learnt, one evaluation metric is how much it improves a NLP task. We choose the Chinese Word Segmentation task as a standard benchmark to examine their efficiency.
One prevailing approach to CWS is casting it to character based sequence tagging problem, where our representations can be applied. A commonly used tagging set is $\mathcal{T} =  \{B, M, E, S\}$, representing the \textbf{b}egin, \textbf{m}iddle, \textbf{e}nd of a word, or \textbf{s}ingle character forming a word.

Given a sequence $\mathbf{X}$ consisted of $n$ features as $\mathbf{X} = (\mathbf{x}_1, \mathbf{x}_2, \ldots, \mathbf{x}_n)$, the goal of sequence tagging based CWS is to find the most possible tags $\mathbf{Y}^* = \{\mathbf{y}_1^*, \dots, \mathbf{y}_n^*\}$:
\begin{equation}
\mathbf{Y}^* = \argmax_{\mathbf{Y} \in \mathcal{T}^n} p (\mathbf{Y} | \mathbf{X}), \label{eq:cws-argmax}
\end{equation}
where $\mathcal{T}=\{B, M, E, S\}$.

Since tagging set restricts the order of adjacent tags, we model them jointly using a conditional random field, mostly following the architecture proposed by Lample et al.\cite{Lample:2016vz}, via stacking two LSTMs with a CRF layer on top of them.

\subsection{Radical LSTM Layer: Character Composition from Radicals}

In this section, we'll review RNN with Bi-LSTM extension briefly, before introducing our character composition network.

\paragraph{LSTM}

Long Short-Term Memory Networks (LSTMs) \cite{hochreiter1997long} are extensions of Recurrent Neural Networks (RNNs). They are designed to combat gradient vanishing issue via incorporating a memory-cell which enables long-range dependencies capturing.

\paragraph{Bi-LSTM}

One LSTM can only produce the representation $\overrightarrow{\mathbf{h}_t}$ of the left context at every character $t$. To incorporate a representation of the right context $\overleftarrow{\mathbf{h}_t}$, a second LSTM which reads the same sequence in reverse order is used. Pair of this forward and backward LSTM is called bidirectional LSTM (Bi-LSTM)~\cite{Graves:2005kt} in literature. By concatenating its left and right context representations, the final representation is produced as $\mathbf{h}_{t} = [\overrightarrow{\mathbf{h}_{t}} ; \overleftarrow{\mathbf{h}_{t}}]$.

We apply a Bi-LSTM to compose character embeddings from radical embeddings in both directions. The raw character is inserted as the first radical, and the semantic component is appended as the last radical. The motivation behind this trick is to make use of LSTM's bias phenomena. In practice, LSTMs usually tend to be biased towards the most recent inputs of the sequence, thus the first one or last one depends on its direction.

\begin{figure}
\centering
\includegraphics[scale=0.15]{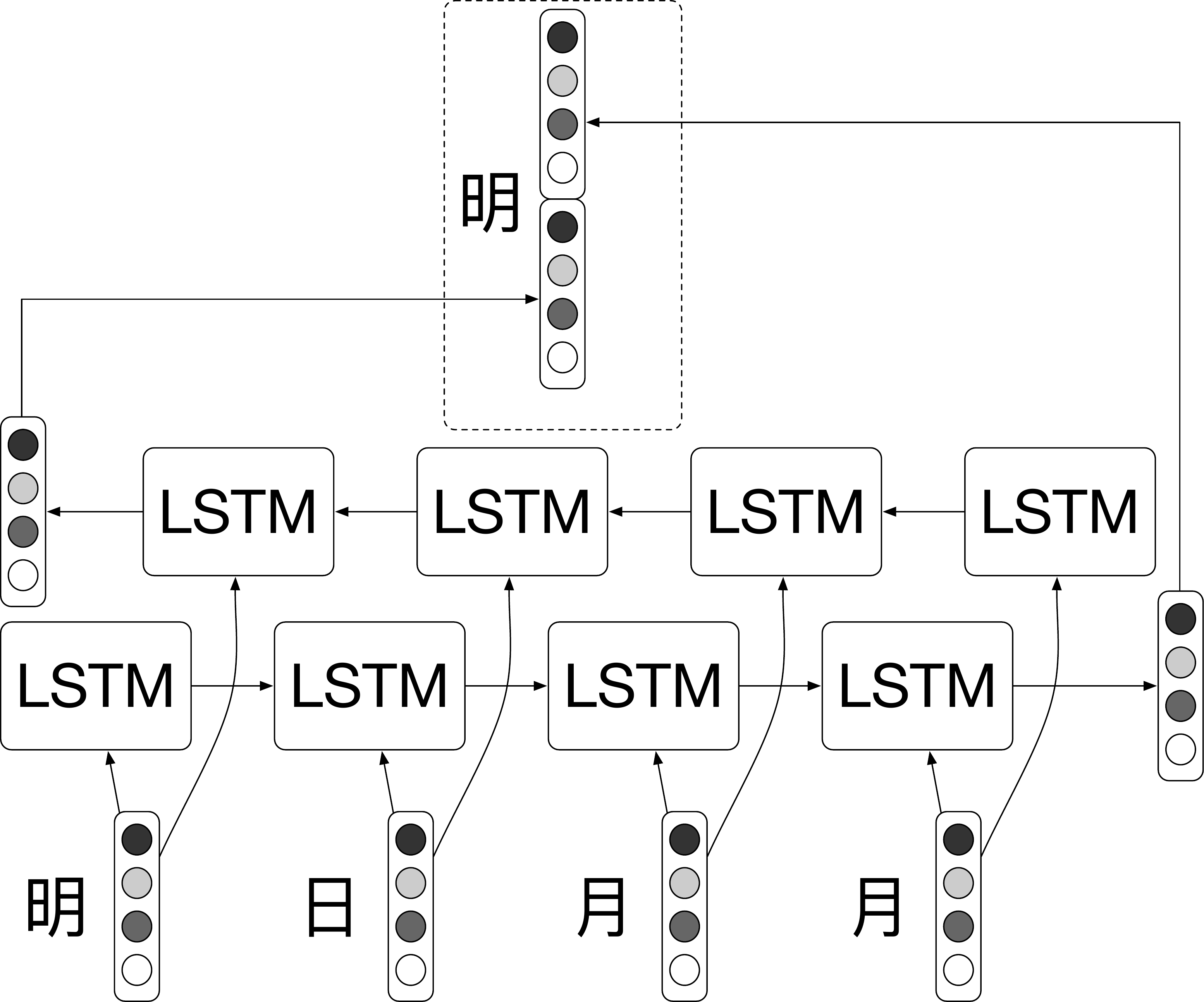}
\caption{Radical LSTM Layer -- composition of character representation from radicals}
\label{radical-lstm}
\vspace{-17pt}
\end{figure}

As illustrated in Figure \ref{radical-lstm}, the character \begin{CJK}{UTF8}{gbsn}明\end{CJK} (bright) has the radical list of \begin{CJK}{UTF8}{gbsn}日\end{CJK} (sun) and \begin{CJK}{UTF8}{gbsn}月\end{CJK} (moon) with its raw form and duplicated semantic radical. Its compositional representation $\mathbf{h}_{i}^{r} \in \mathbb{R}^{2k}$ is agglomerated via a Bi-LSTM from these radical embeddings, where $k$ is the dimension of radical embeddings.

\subsection{Character Bi-LSTM Layer: Context Capturing}

Once compositional character representation $\mathbf{h}_{i}^{r}$ is synthesized, the contextual representation $\mathbf{h}_{t}^c \in \mathbb{R}^{2d}$ at every character $t$ in input sentence can be agglomerated by a second Bi-LSTM. The dimension $d$ is a flexible hyper-parameter, which will be explored in later experiments.

\begin{figure}
\centering
\includegraphics[scale=0.15]{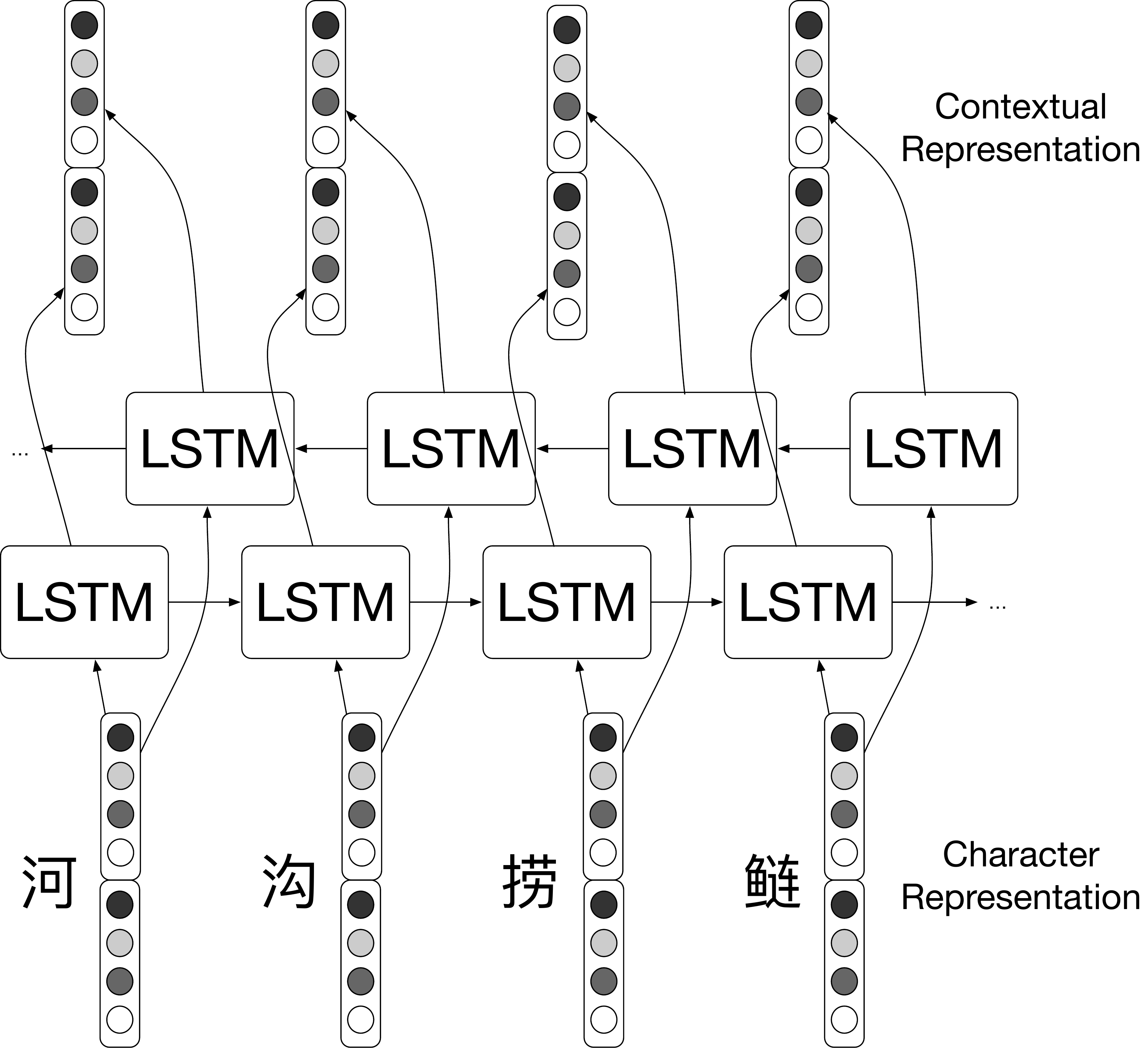}
\caption{Character LSTM Layer -- capture contextual representation}
\label{character-lstm}
\vspace{-17pt}
\end{figure}

Our architecture for contextual feature capturing is shown in Figure \ref{character-lstm}. This contextual feature vector contains the meaning of a character, its radicals and its context.

\subsection{CRF Layer: Tagging Inference}

We employed a Conditional Random Fields(CRF) \cite{Lafferty2001Conditional} layer as the inference layer. As first order linear chain CRFs only model bigram interactions between output tags, so the maximum of a posteriori sequence $\mathbf{Y}^*$ in Eq.~\ref{eq:cws-argmax} can be computed using dynamic programming, both in training and decoding phase. The training goal is to maximize the log-probability of the gold tag sequence.

\section{Experiments}

We conducted various experiments to verify the following questions:

\begin{enumerate}
\item Does radical embedding enhance character embedding in pre-training phase?
  \item Whether radical embedding helps character embedding in training phase and test phase (by using character embedding solely or using them both)?
  \item Can radical embedding replace character embedding (by using radical embedding only)?
  \item Should we tie up two level embeddings?
\end{enumerate}

\subsection{Datasets}

To explore these questions, we experimented on the 4 prevalent CWS benchmark datasets from SIGHAN2005 \cite{emerson_second_2005}. Following conventions, the last 10\% sentences of training set are used as development set.

\subsection{Radical Decomposition}

We obtained radical lists of character from the \textit{online Xinhua Dictionary}\footnote{\url{http://tool.httpcn.com/Zi/}}, which are included in our open-source project.

\subsection{Pre-training}

Previous work have shown that pre-trained embeddings on large unlabeled corpus can improve performance. It usually involves lots of efforts to preprocess those corpus. Here we presented a novel solution.

The corpus used is Chinese Wikipedia of July 2017. Unlike most approaches, we don't perform Traditional Chinese to Simplified Chinese conversion. Our radical decomposition is sufficient of associate character to its similar variants. Not only traditional-simplified character pairs, those with similar radical decompositions will also share similar vectorial representations.

Further, instead of the commonly used word2vec \cite{Mikolov:2013wc}, we utilized fastText\footnote{\url{https://github.com/facebookresearch/fastText} With tiny modification to output $n$-gram vectors.} \cite{2016arXiv160704606B} to train character embeddings and radical embeddings jointly. We applied SG model, $100$ dimension, and set both maximum and minimal $n$-gram length to $1$, as the radical takes only one token.

\subsection{Final Results on SIGHAN bakeoff 2005}

Our baseline model is Bi-LSTM-CRF trained on each datasets only with pre-trained character embedding (the conventional word2vec), no sub-character enhancement, no radical embeddings. Then we improved it with sub-character information, adding radical embeddings, tying two level embeddings up. The final results are shown in Table~\ref{bakeoff-result}.

\begin{table}
\centering
\caption{
Comparison with previous state-of-the-art models of results on all four Bakeoff-2005 datasets.
}
\begin{tabular}{c|cccc}
\hline
Models&PKU & MSR & CityU & AS\\
\hline
Tseng et al. \cite{Tseng2005A} & 95.0 & 96.4 & - & - \\
Zhang and Clark \cite{zhang_chinese_2007} & 95.0 & 96.4 & - & - \\
Sun et al. \cite{sun2009a}& 95.2&97.3&-&-\\
Sun et al. \cite{sun2012fast}& 95.4&\textbf{97.4}&-&-\\
Pei et al. \cite{Pei:2014vx}       &95.2&97.2&-&-\\
Chen et al. \cite{2017arXiv170407556C}                    &94.3&96.0&95.6&94.8\\
Cai et al. \cite{2017arXiv170407047C}$^{\diamondsuit}$&\textbf{95.8}  & 97.1 & 95.6 & 95.3\\

\hline
baseline &94.6 & 96.0 & 94.7 & 94.8\\
+subchar &95.0 & 96.0 & 94.9 & 94.9\\
+radical &94.6 & 96.7 & 95.3 & 95.2\\
+radical -char&94.4 & 96.5 & 95.0 & 95.1\\
+radical +tie&94.8 & 96.8 & 95.3 &  95.1\\
+radical +tie +bigram&95.3 & \textbf{97.4} & \textbf{95.9} & \textbf{95.7}\\
\hline
\end{tabular}
\vspace{-16pt}
\label{bakeoff-result}
\end{table}

All experiments are conducted with standard Bakeoff scoring program\footnote{\url{http://www.sighan.org/bakeoff2003/score} This script rounds a score to one digit.} calculating precision, recall, and $\text{F}_1$-score. Note that results with $\diamondsuit$ expurgated long words in test set.

\subsection{Model Analysis}

Sub-character information enhances character embeddings. Previous work showed pre-trained character embeddings can improve performance. Our experiment showed with sub-character information (+subchar), performance can be further improved compared to no sub-character enhancement (baseline). By simply replacing the conventional word2vec embeddings to radical aware embeddings, the score can benefit an improvement as much as $0.4\%$.

Radical embeddings collaborate well with character embeddings. By building compositional embeddings from radical level (+radical), performance increased by up to $0.7\%$ in comparison with model (baseline) on MSR dataset. But we also notice that: 1) On small dataset such as PKU, radical embeddings cause tiny performance drop. 2) With the additional bigram feature, performance can be further increased as much as $0.6\%$.

Radical embeddings can't fully replace character embeddings. Without character embeddings but use radical embeddings solely (+radical -char), performance drops a little ($0.1\%$ to $0.3\%$) compared to the model with character embeddings (+radical).

Tying two level embeddings up is a good idea. By tying radical embeddings and character embeddings together (+radical +tie), the raw feature is unified into the same vector space, knowledge is transferred between two levels, and performance is boosted up to $0.2\%$.

\section{Conclusions and Future Work}

In this paper, we proposed a novel neural network architecture with dedicated pre-training techniques to learn character and sub-character representations jointly. As an concrete application example, we unified Simplified and Traditional Chinese characters through sub-character or radical embeddings. We have utilized a practical way to train radical and character embeddings jointly. Our experiments showed that sub-character information can enhance character representations for a pictographic language like Chinese. By using both level embeddings and tying them up, our model has gained the most benefit and surpassed previous single criterial CWS systems on 3 datasets.

Our radical embeddings framework can be applied to extensive NLP tasks like POS-tagging and Named Entity Recognition (NER) for various hieroglyphic languages. These tasks will benefit from deeper level of semantic representations encoded with more linguistic knowledge.

\bibliographystyle{IEEEtran}%
\bibliography{references}%
\end{document}